\documentclass{article}
\usepackage{mlspconf}

\usepackage{bbm}
\usepackage{array}
\usepackage{dcolumn}
\usepackage{epsfig}
\usepackage[intlimits]{amsmath}
\usepackage{amsmath, amsfonts, yhmath, bm}
\usepackage{amssymb}
\usepackage{amsthm}
\usepackage{psfrag}
\usepackage{color,soul}
\usepackage[dvipsnames]{xcolor}
\usepackage[normalem]{ulem}
\usepackage{enumerate}
\usepackage{stackengine}
\usepackage[noadjust]{cite}
\usepackage{graphicx}
\usepackage{caption}
\usepackage{subcaption}
\usepackage[font=footnotesize]{subcaption}
\usepackage{multirow}
\usepackage[font=footnotesize]{caption}
\usepackage{etoolbox}
\usepackage{tcolorbox}
\usepackage{float}
\usepackage{dsfont}
\usepackage{tikz}
\usetikzlibrary{arrows}
\usepackage{eucal}
\usepackage[implicit=false]{hyperref}
\hypersetup{
    colorlinks=false,
    pdfborder={0 0 0},
}

\newcommand {\myvec}[1] {{\mbox{\boldmath $#1$}}}
\newcommand {\mymat}[1]  {{\mbox{\boldmath $#1$}}}
\newcommand {\myten}[1]  {{\mathbfcal{#1}}}

\newcommand*{\myfontb}{\fontfamily{lmr}\selectfont}

\DeclareMathAlphabet      {\mathbfit}{OML}{cmm}{b}{it}
\DeclareMathAlphabet	  {\mathbfcal}{OMS}{cmsy}{b}{n}

\newcommand {\mS} {\mymat{S}}
\newcommand {\A} {\mymat{A}}

\newcommand {\G} {\mymat{G}}

\newcommand {\bS} {\mybar{\mS}}
\newcommand {\bX} {\mybar{\X}}

\newcommand {\B} {\mymat{B}}

\newcommand {\D} {\mymat{D}}

\renewcommand {\H} {\mymat{H}}

\newcommand {\Q} {\mymat{Q}}

\newcommand {\X} {\mymat{X}}

\newcommand {\ue} {\myvec{e}}

\newcommand {\ub} {\myvec{b}}

\newcommand {\ux} {\myvec{x}}
\newcommand {\ud} {\myvec{d}}

\newcommand {\up} {\myvec{p}}

\newcommand {\uv} {\myvec{v}}

\newcommand {\us} {\myvec{s}}

\newcommand {\uw} {\myvec{w}}
\newcommand {\uz} {\myvec{z}}
\newcommand {\uh} {\myvec{h}}

\newcommand {\Rset} {\mathbb{R}}
\newcommand {\Cset} {\mathbb{C}}
\newcommand {\Zset} {\mathbb{Z}}
\newcommand {\Eset} {\mathbb{E}}

\newcommand {\tps} {\mathrm{T}}
\DeclareMathOperator{\Diag}{Diag}

\newcommand\norm[1]{\left\lVert#1\right\rVert}
\newcommand {\sbar} {\mybar{s}}
\newcommand {\xbar} {\mybar{x}}
\newcommand {\vbar} {\mybar{v}}

\newcommand {\buv} {\mybar{\uv}}
\newcommand {\bus} {\mybar{\us}}

\newcommand {\bux} {\mybar{\ux}}
\newcommand {\her} {\mathrm{H}}
\newcommand {\hup} {\widehat{\up}}
\newcommand {\MFP} {\text{\tiny MFP}}

\newcommand {\SBL} {\text{\tiny SBL}}

\let\oldbibliography\thebibliography
\renewcommand{\thebibliography}[1]{%
  \oldbibliography{#1}%
  \setlength{\itemsep}{0pt}%
}
\makeatletter
\newsavebox\myboxA
\newsavebox\myboxB
\newlength\mylenA

\newcommand*\mybar[2][0.75]{%
	\sbox{\myboxA}{$\m@th#2$}%
	\setbox\myboxB\null% Phantom box
	\ht\myboxB=\ht\myboxA%
	\dp\myboxB=\dp\myboxA%
	\wd\myboxB=#1\wd\myboxA% Scale phantom
	\sbox\myboxB{$\m@th\overline{\copy\myboxB}$}%  Overlined phantom
	\setlength\mylenA{\the\wd\myboxA}%   calc width diff
	\addtolength\mylenA{-\the\wd\myboxB}%
	\ifdim\wd\myboxB<\wd\myboxA%
	\rlap{\hskip 0.5\mylenA\usebox\myboxB}{\usebox\myboxA}%
	\else
	\hskip -0.5\mylenA\rlap{\usebox\myboxA}{\hskip 0.5\mylenA\usebox\myboxB}%
	\fi}
\makeatother
\newtheorem{prop}{Proposition}

\DeclareMathOperator*{\argmin}{argmin}
\DeclareMathOperator*{\argmax}{argmax}

\title{Direct Localization in Underwater Acoustics via Convolutional Neural Networks: A Data-Driven Approach}
\name{Amir Weiss, Toros Arikan and Gregory W. Wornell\thanks{Code available at \scriptsize {\url{https://www.weissamir.com/project/DLOC}.}}\thanks{This work was supported, in part, by ONR under Grant No.\ N00014-19-1-2665, and NSF under Grant No.\ CCF-1816209.}}
\address{Massachusetts Institute of Technology}

\begin{document}
%\ninept

\maketitle

\begin{abstract}
Direct localization (DLOC) methods, which use the observed data to localize a source at an unknown position in a one-step procedure, generally outperform their indirect two-step counterparts (e.g., using time-difference of arrivals). However, underwater acoustic DLOC methods require prior knowledge of the environment, and are computationally costly, hence slow. We propose, what is to the best of our knowledge, the first data-driven DLOC method. Inspired by classical and contemporary optimal model-based DLOC solutions, and leveraging the capabilities of convolutional neural networks (CNNs), we devise a holistic CNN-based solution. Our method includes a specifically-tailored input structure, architecture, loss function, and a progressive training procedure, which are of independent interest in the broader context of machine learning. We demonstrate that our method outperforms attractive alternatives, and asymptotically matches the performance of an oracle optimal model-based solution.
% Direct localization (DLOC) methods, which use the observed data to localize a source at an unknown position in a one-step procedure, are known to generally attain higher accuracies than their indirect two-step counterparts (e.g., using time-difference of arrivals). However, in the context of underwater acoustics, DLOC approaches require prior knowledge of the environment, and are computationally costly, hence slow. We propose, what is to the best of our knowledge, the first data-driven DLOC method. Inspired by classical and contemporary optimal model-based DLOC solutions, and leveraging the capabilities of convolutional neural networks (CNNs), we devise a holistic CNN-based solution. Our method includes a specifically tailored input structure, architecture, loss function, and a progressive training procedure, which are likely to be of independent interest in the broader context of machine learning. We demonstrate that our method outperforms attractive alternatives, and asymptotically matches the performance of an oracle optimal model-based solution.
\end{abstract}
\vspace{-0.3cm}
\section{Introduction}\label{sec:intro}
\vspace{-0.2cm}
Underwater acoustic localization (UAL) is an important and challenging problem that arises in a wide range of emerging applications \cite{waterston2019ocean}. As such, it has been extensively addressed, and a host of methods---for the most part model-based---have been proposed for different operational environments  \cite{tan2011survey}.

Due to the unprecedented success of deep neural networks (DNNs) in various domains in recent years, the UAL problem had also been approached in a data-driven manner \cite{bianco2019machine}. Motivated by the sensational achievements in classification using CNNs in the visual domain, some methods have been proposed that use such a classification-based methodology for UAL \cite{niu2019deep,testolin2019underwater,gong2020machine,chen2021model}. However, this type of framework may not be a well-suited approach to working with 1-dimensional (non-speech) acoustic signals, and a CNN trained for regression may be a more natural fit. Other regression-type solutions \cite{lefort2017direct,wang2018underwater,qin2020underwater} were not necessarily tailored to the UAL problem, in the sense that some architectural choices, such as the loss function or training procedure, were made according to standard, ``off-the-shelf" ones, which are not necessarily suitable to this challenging domain.

A particularly attractive approach for UAL is \emph{direct} localization (DLOC) \cite{weiss2004direct,wang2020direct}, which is theoretically superior (in terms of accuracy) to indirect methods, wherein first, some statistics of the data (e.g., time-differences of arrivals (TDOAs)) are estimated, and only then is localization performed based on the estimated statistics. DLOC methods require model characterization of the environment of operation, and typically have a high computational cost. To the best of our knowledge, a data-driven DLOC method, with a specifically-crafted DNN, has not been proposed to date.

In this work, we propose such a DLOC solution. By analyzing DLOC optimal model-based methods, we identify key ingredients in their structures, based on which we devise the input structure, architecture, loss function, and a progressive training procedure. Our main contributions, which are potentially of independent interest beyond the UAL context, are: (i) A novel CNN-based DLOC solution, which, to the best of our knowledge, is the first of its kind; (ii) An extended, model-based semi-blind localization (SBL, \cite{weiss2021semi}) solution for UAL; (iii) A progressive training procedure for a model aiming for the joint estimation of several parameters; and (iv) A suitable loss function for learning to estimate a periodic parameter.

\vspace{-0.35cm}
\section{Problem Formulation}\label{sec:problemformulation}
\vspace{-0.2cm}
Consider $L$ spatially-diverse, time-synchronized receivers at known locations, each consisting of a single omni-directional hydrophone. Furthermore, consider the presence of an unknown signal, emitted from a source whose unknown position is denoted by the vector of coordinates $\up\in\Rset^{3\times 1}$. We assume that the source is static during the observation time interval, and is located sufficiently far from all $L$ receivers to permit a planar wavefront (far-field) approximation, which becomes reasonably accurate in shallow waters at high frequencies.

In such an environment, the underwater acoustic medium gives rise to a rich multipath channel, which can be approximately described using ray propagation \cite{etter2018underwater}. In this model, the acoustic wave that is emitted from the source and measured at the the receiver is represented as a sum of (possibly infinitely many) rays, each propagating according to physical laws governed by the environment's characteristics (e.g., salinity, temperature, bathymetry, etc.). We assume the existence of a parametric representation of the environment, and denote by $\mathcal{E}$ the set of the environmental parameters. Importantly, each ray travels a certain distance from the source to the receiver, and arrives after a time-delay $\tau(\up,\mathcal{E})$, which depends on the source's location $\up$ and the environment $\mathcal{E}$. Note that in the underwater acoustic environment, the relation between $\tau(\up,\mathcal{E})$ and the distance traveled by the corresponding ray is not necessarily simple, and in particular, linear, as is the case for isovelocity environments.

Formally, the sampled, baseband-converted signal from the $\ell$-th receiver is given by
\begin{equation}\label{modelequation}
\begin{gathered}
x_{\ell}[n]=\sum_{r=1}^{R}b_{r\ell}s_{r\ell}[n]+v_{\ell}[n]\triangleq\us_{\ell}^{\tps}[n]\ub_{\ell}+v_{\ell}[n]\in\Cset,\\
n=1,\ldots,N, \;\forall \ell\in\{1,\ldots,L\},
\end{gathered}
\end{equation}
where we have defined $\us_{\ell}[n]\hspace{-0.05cm}=\hspace{-0.05cm}[s_{1\ell}[n] \cdots s_{R\ell}[n]]^{\tps}\hspace{-0.05cm}\in\hspace{-0.05cm}\Cset^{R\times1}$ and $\ub_{\ell}=[b_{1\ell} \cdots b_{R\ell}]^{\tps}\in\Cset^{R\times1}$, using the notations:
{\begin{enumerate}
    \setlength{\itemsep}{0pt}
	\item $b_{r\ell}\in\Cset$ as the unknown attenuation coefficient from the source to the $\ell$-th sensor associated with the $r$-th signal component (line-of-sight (LOS) or other non-LOS (NLOS) reflections);
	\item $s_{r\ell}[n]\triangleq \left.s\left(t-\tau_{r\ell}(\up,\mathcal{E})\right)\right\vert_{t=nT_s}\in\Cset$ as the sampled $r$-th component of the unknown emitted signal's waveform at the $\ell$-th sensor, where $s\left(t-\tau_{r\ell}(\up,\mathcal{E})\right)$ is the analog, continuous-time waveform delayed by $\tau_{r\ell}(\up,\mathcal{E})$, and $T_s$ is the (known) sampling period; and
	\item $v_{\ell}[n]\in\Cset$ as the additive noise at the $\ell$-th receiver, representing the overall contributions of internal receiver noise and ocean ambient noise, modeled as a zero-mean random process with an unknown variance $\sigma_{v_{\ell}}^2$.
\end{enumerate}}

Applying the normalized DFT\footnote{$\mybar{\uz}$ denotes the normalized discrete Fourier transform (DFT) of $\uz$.} to \eqref{modelequation} yields the equivalent frequency-domain representation for all $\ell\in\{1,\ldots,L\}$,
\begin{equation}\label{modelequationfreq}
\begin{aligned}
\hspace{-0.2cm}\xbar_{\ell}[k]&=\sum_{r=1}^{R}b_{r\ell}\sbar[k]e^{-\jmath\omega_k\tau_{r\ell}({\text{\boldmath $p$}},\mathcal{E})}+\vbar_{\ell}[k]\\
&=\sbar[k]\hspace{-0.05cm}\cdot\hspace{-0.05cm}\underbrace{\ud_{\ell}^{\her}[k]\ub_{\ell}}+\vbar_{\ell}[k]=\sbar[k]\hspace{-0.05cm}\cdot\hspace{-0.05cm}\underbrace{\bar{h}_{\ell}[k]}+\vbar_{\ell}[k]\in\Cset,
\end{aligned}
\end{equation}
where we have defined $\bar{h}_{\ell}[k]\triangleq\ud_{\ell}^{\her}[k]\ub_{\ell}\in\Cset$, and
\begin{equation*}\label{dft_freq_vec}
\begin{gathered}
\ud_{\ell}[k]\triangleq[e^{-\jmath\omega_k\tau_{1\ell}({\text{\boldmath $p$}},\mathcal{E})} \cdots e^{-\jmath\omega_k\tau_{R\ell}({\text{\boldmath $p$}},\mathcal{E})}]^{\her}\in\Cset^{R\times 1},\\
\omega_k\hspace{-0.05cm}\triangleq\hspace{-0.05cm}\frac{2\pi(k-1)}{NT_s}\in\Rset_+, \quad k=1,\ldots,N.
\end{gathered}
\end{equation*}
As shorthand notation, we further define
\begin{equation*}%\label{initialdefinitionsofsignals}
\begin{aligned}
\bux_{\ell}&\hspace{-0.025cm}\triangleq\hspace{-0.025cm}\left[\mybar{x}_{\ell}[1] \cdots \mybar{x}_{\ell}[N]\right]^{\tps}, \bus\triangleq \left[\sbar[1] \cdots \sbar[N]\right]^{\tps},\\
\buv_{\ell}&\hspace{-0.025cm}\triangleq\hspace{-0.025cm} \left[\mybar{v}_{\ell}[1] \cdots \mybar{v}_{\ell}[N]\right]^{\tps}\hspace{-0.05cm}, \D_{\ell}\triangleq\left[\ud_{\ell}[1] \cdots \ud_{\ell}[N]\right]^{\tps}\hspace{-0.05cm}\in\Cset^{N\times R},\\
\bX_{\ell}&\hspace{-0.025cm}\triangleq\hspace{-0.025cm}\Diag(\bux_{\ell}), \bS\triangleq\Diag(\bus), \H_{\ell}(\up,\mathcal{E})\triangleq\Diag\left(\D_{\ell}\ub_{\ell}\right),
\end{aligned}
\end{equation*}
where $\Diag(\cdot)$ forms a diagonal matrix from its vector argument. Note that $\H_{\ell}(\up,\mathcal{E})$ is generally a nonlinear function of the unknown source position $\up$ and the environmental parameters $\mathcal{E}$, e.g., as seen from by the definition of $\ud_{\ell}[k]$. With this notation, we may now write \eqref{modelequationfreq} compactly as
\begin{equation}\label{signalmodelfreqcompact}
\bux_{\ell}=\H_{\ell}(\up,\mathcal{E})\bus+\buv_{\ell}\in\Cset^{N\times 1}, \;\forall\ell\in\{1,\ldots,L\}.
\end{equation}

Thus, the localization problem is formulated as follows:\vspace{0.1cm}
%\tcbset{colframe=gray!95!blue,size=small,width=0.49\textwidth,arc=2.1mm,outer arc=1mm}
%\begin{tcolorbox}[upperbox=visible,colback=white]
%\noindent\textbf{Problem:} 
{\myfontb\emph{Given the observations $\left\{\bux_{\ell}\in\Cset^{N\times 1}\right\}_{\ell=1}^{L}$ of the signal as per model \eqref{signalmodelfreqcompact}, localize the source. Specifically, design an estimator $\hup$ of $\up$, so as to minimize $\Eset\left[\|\hup-\up\|_2^2\right]$.}}\vspace{0.1cm}
%\end{tcolorbox}

In this paper, we assume that we do not have access to precise knowledge of the environment $\mathcal{E}$ nor the underlying propagation model, from which $\H_{\ell}(\up,\mathcal{E})$ can be computed for any $\up$. Rather, we assume that a dataset of signals with labeled source positions $\mathcal{D}^{(J)}_{\text{train}}\triangleq\{(\bux^{(i)}_{1},\ldots,\bux^{(i)}_{L},\up^{(i)})\}_{i=1}^{J}$ of size $J$ is available. Such a dataset can be obtained by taking measurements using a deployed system, or by generating synthetic signals, possibly using simulators such as Bellhop \cite{porter2011bellhop}, and combining them with recorded ocean ambient noise, which is readily accessible, e.g., \cite{hodgkiss2012kauai,menze2017influence}.

%We emphasize that although we are interested solely in $\up$, the channel parameters $\{\ub_{\ell}\}$ and the DFT coefficients $\bus$ of the emitted waveform are unknown as well.
\vspace{-0.3cm}
\section{Optimal Model-Based Solutions}\label{sec:SBLthreeraymodel}
\vspace{-0.2cm}
To motivate our proposed approach and gain theoretically-justified intuition into the implementation and architectural choices, it is instructive to first present two model-based solutions, which are optimal, each for its respective model, under slightly different assumptions.

\vspace{-0.3cm}
\subsection{Matched Field Processing}\label{subsec:mfpsolution}
\vspace{-0.2cm}
The key assumption of Matched Field Processing (MFP) \cite{baggeroer1993overview} approaches is that, for a given hypothesized source location $\up$, the channel response $\H_{\ell}(\up,\mathcal{E})$ is \emph{fully} predictable, known,\footnote{Otherwise, infeasible high-dimensional optimization is required.} and computable. In many practical cases, this assumption is not a realistic one. For example, the attenuation coefficients associated with the bottom reflections generally depend on the angle of incidence, the sediment type at the area of operation (e.g., sand, silt, clay, etc.) and various other factors.% Moreover, $\{\ub_{\ell}\}$ can vary (even if slowly) with time at a given environment. That is, even if they are constant during a given observation period, they may have different constant values during another observation period at a different time.

Nevertheless, assuming that $\mathcal{E}$ is perfectly known, the MFP solution serves as a theoretical benchmark, since it is the maximum likelihood estimator (MLE) under the additive white Gaussian noise model, and hence is asymptotically efficient (in the statistical sense). It can be shown that after simplification (see, e.g., \cite{weiss2021semi}), the MFP solution is given by
% Accordingly, it is defined as 
% \begin{equation}\label{MFPoptimmization}
% \hup_{\MFP} \triangleq \underset{\text{\boldmath$p$}\in\Rset^{3\times1}}{\argmin} \; 
% \left\{\min_{\text{\boldmath$\bar{s}$}\in\Cset^{N\times 1}}\sum_{\ell=1}^{L}\norm{\bux_{\ell}-\H_{\ell}(\up,\mathcal{E})\bus}_2^2\right\},
% \end{equation}
% and is given more explicitly by the simplified expression
\begin{equation}\label{MFPsimpleform}
\hup_{\MFP}=\underset{\text{\boldmath$p$}\in\Rset^{3\times1}}{\argmax} \;  \sum_{k=1}^{N}\frac{\left|\bux[k]^{\her}\mybar{\uh}_k(\up,\mathcal{E})\right|^2}{\|\mybar{\uh}_k(\up,\mathcal{E})\|_2^2},
\end{equation}
where we have defined, for every $k$-th DFT component,
\begin{align*}%\label{newdeffreqvectors}
\bux[k]&\triangleq\left[\bar{x}_1[k]\,\cdots\,\bar{x}_L[k]\right]^{\tps}\in\Cset^{L\times 1},\\
\mybar{\uh}_k(\up,\mathcal{E})&\triangleq\left[\bar{h}_{1}[k]\;\cdots\;\bar{h}_{L}[k]\right]^{\tps}\in\Cset^{L\times 1}.
\end{align*}
As mentioned above, MFP requires the precise knowledge of the channel impulse response. When this knowledge is not available, \eqref{MFPsimpleform} can be thought of as an oracle MLE.

\vspace{-0.3cm}
\subsection{Semi-Blind Localization}\label{subsec:sblsolution}
\vspace{-0.2cm}
Unlike the MFP framework, the formulation of the semi-blind localization (SBL) problem \cite{weiss2021semi} considers the channel attenuation coefficients $\{\ub_{\ell}\}$ to be arbitrary (deterministic) unknown parameters, rather than as functions of $\up$ and/or $\mathcal{E}$. Thus, by intentionally ignoring the functional structure of $\{\ub_{\ell}\}$, additional (possibly redundant) nuisance parameters are introduced, and must be (implicitly) estimated in order to estimate $\up$. It turns out that this redundancy pays off in terms of robustness, and the SBL estimator is more resilient than other methods to some  unmodeled environmental structures, such as unknown occluders \cite[Sec.~VI]{weiss2021semi}.

We now present an extended version of the SBL method, which had been proposed for a three-ray model that takes into account the LOS component, and the primary NLOS surface and bottom reflections. This extension is generally of independent interest for UAL, beyond the main contributions of this paper, to be presented in the next section.

For the general ray propagation model \eqref{modelequation}, possibly with more than three rays and not necessarily for an isovelocity environment, we define the (extended) SBL estimator as
\begin{equation}\label{MLEoptimmization}
\hup_{\SBL} \triangleq \underset{\text{\boldmath$p$}\in\Rset^{3\times1}}{\argmin} \; \left\{\min_{\substack{\text{\boldmath$\bar{s}$}\in\mathcal{S}_N \\ {\text{\boldmath$B$}}\in\Cset^{R\times L}}} \sum_{\ell=1}^{L}\norm{\bux_{\ell}-\H_{\ell}(\up,\mathcal{E})\bus}_2^2\right\},
\end{equation}
where $\bus\in\mathcal{S}_N\triangleq\{\uz\in\Cset^{N\times1}\colon\|\uz\|_2=1\}$ is used in \eqref{MLEoptimmization} without loss of generality, since both $\bus$ and $\ub_{\ell}$ are unknown,\footnote{Giving rise to an inherent scaling ambiguity: $\bus\cdot\ub_{\ell}=\left(\alpha\bus\right)\cdot\left(\frac{1}{\alpha}\ub_{\ell}\right)$.} and $\B\triangleq\left[\ub_1 \cdots \ub_L\right]\in\Cset^{R\times L}$. We have the following result:
% \begin{prop}[SBL for a general ray model]\label{maxeigenvalueflatspectrum}
\begin{prop}\label{maxeigenvalueflatspectrum}
Assume $\{\D_{\ell}\}_{\ell=1}^L$ are full column rank. For any potential position $\up$, define the data-dependent matrix,
\begin{equation}\label{targetmatrixdef}
\Q(\up,\mathcal{E})\triangleq\sum_{\ell=1}^{L}\bX_{\ell}\D^*_{\ell}\left(\D_{\ell}^{\tps}\D^*_{\ell}\right)^{-1}\left(\bX_{\ell}\D^*_{\ell}\right)^{\her}\in\Cset^{N\times N}.
\end{equation}
Then, for any spectrally flat waveform, i.e., $|\sbar[k]|=\rho$, for all $k\in\{1,\ldots,N\}$ and some $\rho>0$, the SBL estimator is
\begin{equation}\label{SBLestimate}
\hup_{\emph{\SBL}}=\underset{\text{\boldmath$p$}\in\Rset^{3\times1}}{\argmax} \; \lambda_{\max}\left(\Q(\up,\mathcal{E})\right),
\end{equation}
where $\lambda_{\max}(\A)$ denotes the largest eigenvalue of $\A\hspace{-0.05cm}\in\hspace{-0.05cm}\Cset^{N\hspace{-0.05cm}\times\hspace{-0.05cm}N}$.
\end{prop}\vspace{-0.1cm}
\noindent Due to space limitations, the proof is given in \cite{weiss2022mlspsuppmat}.
% \begin{proof}
% We observe that introducing the dependence in the environment parameters of the set $\mathcal{E}$ into the matrices $\{\D_{\ell}\}$ is only through the time-delays $\{\tau_{r\ell}(\up,\mathcal{E})\}$. Further, introducing the $R-3$ additional rays beyond the three corresponding to the LOS and NLOS primary surface and bottom reflections, result only in an increased column dimension, but otherwise does not change its general structure. From this point, the proof continues as the one presented in \cite[Appendix A]{weiss2021semi}.
% \end{proof}

Although \eqref{SBLestimate} holds exactly for spectrally flat signals, it is also a reliable solution for general waveforms \cite[Prop.~3]{weiss2021semi}. However, the complexity of \eqref{SBLestimate} is ${\CMcal{O}(RN)}$, and it can be implemented only when the time-delays $\{\tau_{r\ell}(\up,\mathcal{E})\}$ can be computed for any $\up$. In practice, this is not always possible, since in some cases $\mathcal{E}$ is not fully known. Moreover, even when $\mathcal{E}$ is known, the computation can be nontrivial, and has to be executed for \emph{any} tested hypothesized position $\up$.

Before we present our proposed solution, we point to the following observations regarding the MFP and SBL solutions:
\begin{itemize}\vspace{-0.1cm}
    \itemsep0em
    \item \textit{Sufficient statistic}: Although MFP and SBL differ in their simplified forms \eqref{MFPsimpleform} and \eqref{SBLestimate}, respectively, both require \emph{only} the (same) correlation functions of the observed data. This can be observed by rewriting the numerator of \eqref{MFPsimpleform}, and writing explicitly the statistically-equivalent form of \eqref{targetmatrixdef}---$\widetilde{\Q}(\up,\mathcal{E})$ (given in \cite{weiss2022mlspsuppmat}), due to $\lambda_{\max}\left(\Q(\up,\mathcal{E})\right)=\lambda_{\max}\left(\widetilde{\Q}(\up,\mathcal{E})\right)$---as%Rewriting the numerator of \eqref{MFPsimpleform}, and writing explicitly the statistically-equivalent form of \eqref{targetmatrixdef}---$\widetilde{\Q}(\up,\mathcal{E})$ (given in \cite{weiss2022mlspsuppmat}), due to $\lambda_{\max}\left(\Q(\up,\mathcal{E})\right)=\lambda_{\max}\left(\widetilde{\Q}(\up,\mathcal{E})\right)$---this is observed from
    \begin{gather*}
     \left|\bux[k]^{\her}\mybar{\uh}_k(\up,\mathcal{E})\right|^2=\mybar{\uh}_k(\up,\mathcal{E})^{\her}\underbrace{\bux[k]\bux[k]^{\her}}\mybar{\uh}_k(\up,\mathcal{E}),\\
     \widetilde{\Q}(\up,\mathcal{E})\hspace{-0.05cm}=\hspace{-0.05cm}\begin{bmatrix}
                            \G_{1}^{\her}\underbrace{\bX_{1}^{\her}\bX_{1}}\G_{1} & \hspace{-0.1cm}\cdots\hspace{-0.1cm} & \G_{1}^{\her}\underbrace{\bX_{1}^{\her}\bX_{L}}\G_{L}\\
                            \vdots & \hspace{-0.1cm}\ddots\hspace{-0.1cm} & \vdots\\
                            \G_{L}^{\her}\underbrace{\bX_{L}^{\her}\bX_{1}}\G_{1} & \hspace{-0.1cm}\cdots\hspace{-0.1cm} & \G_{L}^{\her}\underbrace{\bX_{L}^{\her}\bX_{L}}\G_{L}\\
                         \end{bmatrix},\\
     \Longrightarrow\;\; \ue^{\tps}_{\ell_1}\bux[k]\bux[k]^{\her}\ue_{\ell_2}=\ue^{\tps}_{k}\bX_{\ell_1}^{\her}\bX_{\ell_2}\ue_{k},
    \end{gather*}
    where $\{\G_{\ell}\in\Cset^{N\times R}\}_{\ell=1}^L$, defined in \cite{weiss2022mlspsuppmat}, are data-independent, and $\ue_n$ is the $n$-th standard basis vector. It follows that MFP and SBL share \emph{the same} sufficient statistic, only arranged differently. Recall that an element-wise product of a DFT vector with the conjugate of another is equivalent to correlating the corresponding time-domain vectors. Hence, the sufficient statistics for the optimal model-based solutions are the second-order statistics (SOS), i.e., the empirical correlation functions of all $L$ observed signals $\{x_{\ell}[n]\}$.\vspace*{-0.1cm}
    \item \emph{Joint optimization of  orthogonal coordinates}: Although range, inclination and azimuth are algebraically orthogonal, and therefore in some sense decoupled, they are nevertheless \emph{statistically coupled}. Thus, DLOC, i.e., the estimation of $\up$ directly from the data can (and does) lead to improved accuracy \cite{weiss2004direct}.
\end{itemize}

\vspace{-0.45cm}
\section{A Data-Driven CNN-based Approach}\label{sec:DLviaCNN}
\vspace{-0.2cm}
\begin{figure*}[t]
	\centering
	\includegraphics[width=0.9\textwidth]{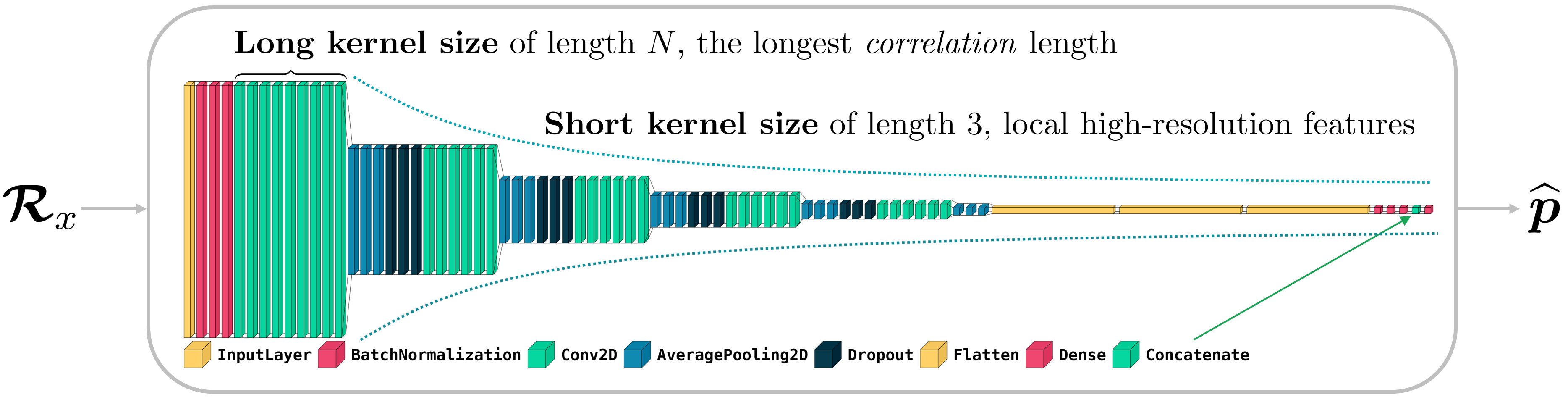}
	\caption{Architecture of the proposed DNN for data-driven DLOC. The model is comprised of three sub-models, which are initially trained individually to estimate range, azimuth and inclination. Each sub-model contains $5$ main building blocks, where each block contains $2$D-convolutional layers, followed by $2$D average pooling. Dropouts are used in the first $4$ blocks. The $3$ sub-models are connected with a dense layer. For more implementation details, see \cite{weiss2022mlspsuppmat}.}\vspace{-0.4cm}
	\label{fig:CNNarchitecture}
\end{figure*}
Bearing in mind the practical challenges faced by UAL model-based approaches, namely the assumption that we have prior knowledge of the environment, and taking into account the prohibitive computational cost of a $3$-dimensional optimization (search for the emitter) in the volume of interest, we propose a different approach. As we show in Section \ref{sec:simulationresults}, our method is able to attain the performance of an \emph{oracle} optimal model-based solution, which has access to perfect knowledge of the impulse response. Furthermore, our solution is exceptionally computationally efficient compared to state-of-the-art alternatives, which typically require some grid-search. Thanks to the data-driven nature of our approach, we bypass the need to analytically describe the physical laws that govern the true underlying propagation model, which are manifested in the resulting impulse responses $\{\H_{\ell}(\up,\mathcal{E})\}$.

We propose to use a CNN-based architecture, that is constructed and trained progressively. Fig.~\ref{fig:CNNarchitecture} illustrates the final structure, where some of the implementation details are provided in the caption; see \cite{weiss2022mlspsuppmat} for the full details. The proposed input structure, chosen architecture and loss function, are designed according to properties shared by the optimal model-based solutions \eqref{MFPsimpleform}, \eqref{SBLestimate}, as described in detail next.

\vspace{-0.3cm}
\subsection{Input: The SOS Tensor}\label{subsec:inputSOS}
\vspace{-0.2cm}
As explained in the Section \ref{sec:SBLthreeraymodel}, the MFP and SBL optimal model-based solutions use (only) all the empirical auto- and cross-correlation functions of the observed signals from all $L$ receivers. Therefore, in order to provide a more succinct version of the data, which is in agreement with what the optimal model-based solutions require, we define the SOS tensor as\footnote{The signals in the sum are zero-padded wherever necessary.}
\begin{gather*}%\label{SOStensor}
    \CMcal{R}_x[\ell_{\text{\tiny corr}}(\ell_1,\ell_2),m,1]\triangleq\Re\left\{ \frac{1}{N}\sum_{n}{x_{\ell_1}[n+m]x^*_{\ell_2}[n]}\right\},\nonumber\\
    \CMcal{R}_x[\ell_{\text{\tiny corr}}(\ell_1,\ell_2),m,2]\triangleq\Im\left\{ \frac{1}{N}\sum_{n}{x_{\ell_1}[n+m]x^*_{\ell_2}[n]}\right\},\nonumber\\
    \Longrightarrow  \; \myten{R}_x\in\Rset^{L_{\text{\tiny corr}}\times (2N-1) \times 2}, \quad L_{\text{\tiny corr}}\triangleq \frac{(L+1)L}{2},
\end{gather*}
and use it as the input to our CNN. Here, $\ell_{\text{\tiny corr}}(\ell_1,\ell_2)$ is a bijective function that maps all the unique pairs of receiver indices to a unique index in $\{1,\ldots,L_{\text{\tiny corr}}\}$ (corresponding to all the auto- and cross-correlation functions, without repetition). 

With this input, it is guaranteed that the model has the data-dependent (but not necessarily prior) information required to (at least) match the performance of the optimal model-based solutions presented in Section \ref{sec:SBLthreeraymodel}. This inspires the choice of $2$D-convolutional layers, as with this choice, the CNN can jointly process the correlation functions per-layer.

\vspace{-0.3cm}
\subsection{Loss Function with Spherical Coordinates}\label{subsec:objectivefunctions}
\vspace{-0.2cm}
As the objective function, we choose to minimize 
\begin{equation}\label{objectivefunction}
\begin{aligned}
    \hspace{-0.13cm}&\CMcal{L}_{\text{\tiny sph}}(\hup(\uw),\up)\hspace{-0.05cm}\triangleq\hspace{-0.05cm}\|\hup(\uw)\hspace{-0.025cm}-\hspace{-0.025cm}\up\|_2^2=\widehat{r}^2(\uw)\hspace{-0.025cm}-\hspace{-0.025cm}r^2\hspace{-0.025cm}-\hspace{-0.025cm}2\widehat{r}(\uw)r\cdot\\
    \hspace{-0.13cm}&\Big[\hspace{-0.025cm}\sin(\widehat{\varphi}(\uw))\sin(\varphi)\cos(\widehat{\theta}(\uw)\hspace{-0.025cm}-\hspace{-0.025cm}\theta)\hspace{-0.05cm}+\hspace{-0.05cm}\cos(\widehat{\varphi}(\uw))\cos(\varphi)\hspace{-0.025cm}\Big],
    % \CMcal{L}_{\text{\tiny sph}}(\hup(\uw),\up)\triangleq&\|\hup(\uw)-\up\|_2^2=\\
    % &\;\;\widehat{r}^2(\uw)-r^2-2\widehat{r}(\uw)r\cdot\\
    % &\Big[\sin(\widehat{\theta}(\uw))\sin(\theta)\cos(\widehat{\varphi}(\uw)-\varphi)\\
    % &\;+\cos(\widehat{\theta}(\uw))\cos(\theta)\Big],
\end{aligned}
\end{equation}
namely, the squared Euclidean norm of the estimation error in \emph{spherical coordinates}, where $r, \theta, \varphi$ are the range, azimuth and inclination, respectively, associated with $\up$ (similarly for $\widehat{r}(\uw), \widehat{\theta}(\uw), \widehat{\varphi}(\uw)$ and $\hup(\uw)$), and $\uw$ is the vector of all the weights of the DNN to be trained. Several considerations lead to this particular choice. 

First, although it is analytically convenient to decouple different coordinates, when training a NN we prefer that the gradient of the loss function with respect to one estimand (e.g., range) would depend on another (e.g., azimuth), so as to \emph{exploit statistical dependencies}. Indeed, it can be seen that the gradient of \eqref{objectivefunction} with respect to $\widehat{r}(\uw)$ is a function of $\widehat{\theta}(\uw)$ and $\widehat{\varphi}(\uw)$. This is not the case for Cartesian coordinates.

Second, in spherical coordinates, two coordinate axes are angles, which are periodic by definition. This way, we create a loss function with \emph{infinitely many globally optimal} points, since the metric \eqref{objectivefunction} is invariant to additions of $2\pi$ (and hence $2\pi\Zset$) to $\widehat{\theta}(\uw)$ and $\widehat{\varphi}(\uw)$. This is also not true for the common, but not necessarily justified, choice of Cartesian coordinates.

While the choices above are inspired by the optimal model-based solutions' structure, there is still a need to account for the optimization procedure for computing an estimate given the data. Based on our empirical observations, training our DNN to directly output an estimate of $\up$ without an educated initialization does not seem to work. This possibly results from a highly non-smooth ``landscape" induced by the objective \eqref{objectivefunction} due to the NLOS components, giving rise to many spurious source locations. We address this difficulty using a progressive training procedure.

\vspace{-0.3cm}
\subsection{A Two-Step Progressive Training Procedure}\label{subsec:progtrain}
\vspace{-0.2cm}
Rather than starting with one model that outputs the location of the source $\up$, we start by training three independent models, for estimating \emph{individually} the range, inclination and azimuth. Although these models may yield sub-optimal estimates of the different coordinates, these are much simpler tasks (known as range and direction-of-arrival estimation). The idea behind progressive training is the following: After each model is trained individually, the models are combined (using the concatenation operator), and then serve as the initial weights for the training process of the DLOC model. Observe that in Fig.~\ref{fig:CNNarchitecture}, except for the input and output layers, all layers appear in multiples of three, which is due to the concatenation of the three sub-models trained to individually estimate $r, \theta, $ and $\varphi$. 

For the range model, we choose $|r-\widehat{r}(\uw_r)|^2$ as the loss function, where $\uw_r$ denotes the vector of weights of the model. As for the azimuth $\theta$, we would like to exploit the inherent periodicity to create multiple globally optimal solutions, as explained in Section \ref{subsec:objectivefunctions}. A suitable loss for the estimation of periodic parameters is the mean cyclic error (MCE),
\begin{align}
{\rm{MCE}}(\widehat{\theta},\theta)&\triangleq 2 - 2\Eset\left[\cos\left(\widehat{\theta}-\theta\right)\right]\in\Rset_+.
\end{align}
Therefore, we choose the empirical MCE loss, defined as
\begin{align*}
{\rm{EMCE}}(\widehat{\theta}(\uw_{\theta}),\theta)\triangleq 2 - 2\cos\left(\widehat{\theta}(\uw_{\theta})-\theta\right),
\end{align*}
where $\uw_{\theta}$ denotes the vector of weights of the model. We use the same loss for the inclination angle, scaled by a factor of $2$, since its range is $[0,\pi]$. To summarize the training process:
\tcbset{colframe=gray!90!blue,size=small,width=0.49\textwidth,halign=flush center,arc=2mm,outer arc=1mm}
\begin{tcolorbox}[upperbox=visible,colback=white,halign=left]
\noindent\textbf{\underline{Phase 1}: Train the \emph{individual} sub-models for $r,\theta, \varphi$.}\vspace{0.1cm}

\noindent\textbf{\underline{Phase 2}: Train the model for DLOC, i.e., $\up$, where the three concatenated ``core branches" are initialized by the weights of the previously trained sub-models.}
\end{tcolorbox}

\vspace{-0.35cm}
\section{Simulation Results}\label{sec:simulationresults}
\vspace{-0.2cm}
% \begin{figure}[t]
%   \centering
%   \includegraphics[width=0.95\linewidth]{geometrical_calculation_new}
%   \caption{A 3-dimensional illustration of the three-ray model geometry, assuming isovelocity (i.e., constant speed of sound) for simplicity of the illustration.}\vspace{-0.3cm}
%   \label{fig:geometry3D}
% \end{figure}
We focus on a special case of the general ray model---the isovelocity three-ray model \cite[Fig.~1]{weiss2021semi}, which considers the three primary, typically most dominant, components of the impulse response: the direct-path LOS and the NLOS surface and bottom reflections.
%A $3$-dimensional illustration of the geometry of this propagation model in an isovelocity environment is given in Fig.~\ref{fig:geometry3D}. 
This model yields a method that already allows for successful localization in nontrivial settings (e.g., complete loss of LOS \cite{weiss2021semi}). Additionally, it enables a comparison to the oracle MFP, so as to evaluate how close the proposed solution can get to the oracle's performance, which constitutes a lower bound. By this, we are able to assess whether our model has the required capacity to learn the environment, the receivers geometry, and the optimal estimation rule \eqref{MFPsimpleform}. We note, however, that our approach can cope with the more complex propagation models of other environments.

We consider a scenario with $L=4$ receivers, in a volume with bottom depth $h=50\,\text{m}$. The locations of the receivers are given in Table \ref{table:positions}. The attenuation coefficients were drawn independently from the complex normal (CN) distribution, such that $\Eset\left[|b_{rl}|^2\right]\hspace{-0.05cm}=\hspace{-0.05cm}1$, with variance $0.1^2$. The speed of sound was set to $c=1500\;\text{m}/\text{s}$, and we used $N=100$.
\begin{table}[t!]
	\centering
	\footnotesize
	\begin{tabular}{|c|c|c|c|l}
		\cline{1-4}
		& $x$ {[}m{]} & $y$ {[}m{]} & $z$ {[}m{]} &  \\ \cline{1-4}
		Receiver 1  &   $150$   &   $-250$ &   $10$    &  \\ \cline{1-4}
		Receiver 2  &   $50$    &   $-250$ &   $15$    &  \\ \cline{1-4}
		Receiver 3  &   $-50$   &   $-250$ &   $20$    &  \\ \cline{1-4}
		Receiver 4  &   $-150$  &   $-250$ &   $25$    &  \\ \cline{1-4}
	\end{tabular}
	\caption{Positions of the four receivers in Cartesian coordinates.}
	\label{table:positions}\vspace{-0.55cm}
\end{table}

We trained our model as described in Section \ref{sec:DLviaCNN} using ``hybrid" received signals $\{\bux^{(i)}_{1},\ldots,\bux^{(i)}_{L}\}$ that were generated according to \eqref{modelequationfreq} with $R\hspace{-0.05cm}=\hspace{-0.05cm}3$. The source's baseband-converted waveform was drawn from the standard CN distribution, i.e., $\sbar[k]\overset{\text{iid}}{\sim}\CMcal{CN}(0,1)$, and the noise realizations were taken from recordings of real ocean ambient noise from the KAM11 experiment \cite{hodgkiss2012kauai}. Thus, the model can potentially learn the underlying statistics of real ocean noise, which is generally not temporally white or Gaussian \cite{stojanovic2009underwater}. The size of the dataset $\mathcal{D}^{(J)}_{\text{train}}$ was $J=10^4$ per SNR level, with twenty equidistant levels on $[-10,30]$ dB, where we set $\forall \ell:\sigma_{v_{\ell}}^2=\sigma_v^2$, and define ${\rm SNR}\triangleq1/\sigma_v^{2}$. For each example, the source position was drawn independently in Cartesian coordinates according to
\begin{equation*}
    \up^{(i)}=[\mathsf{x}^{(i)}\;\mathsf{y}^{(i)}\;\mathsf{z}^{(i)}]^{\tps}:\; \begin{cases}
    \mathsf{x}^{(i)}\sim{\rm Unif}\left(-150,150\right)[m],\\
    \mathsf{y}^{(i)}\sim{\rm Unif}\left(-100,0\right)[m],\\
    \mathsf{z}^{(i)}\sim{\rm Unif}\left(5,45\right)[m].
    \end{cases}
\end{equation*}
More details regarding the training process (optimizer, learning rate, etc.) are given in the supplementary materials \cite{weiss2022mlspsuppmat}.

\begin{figure}[t]
	\includegraphics[width=0.95\linewidth]{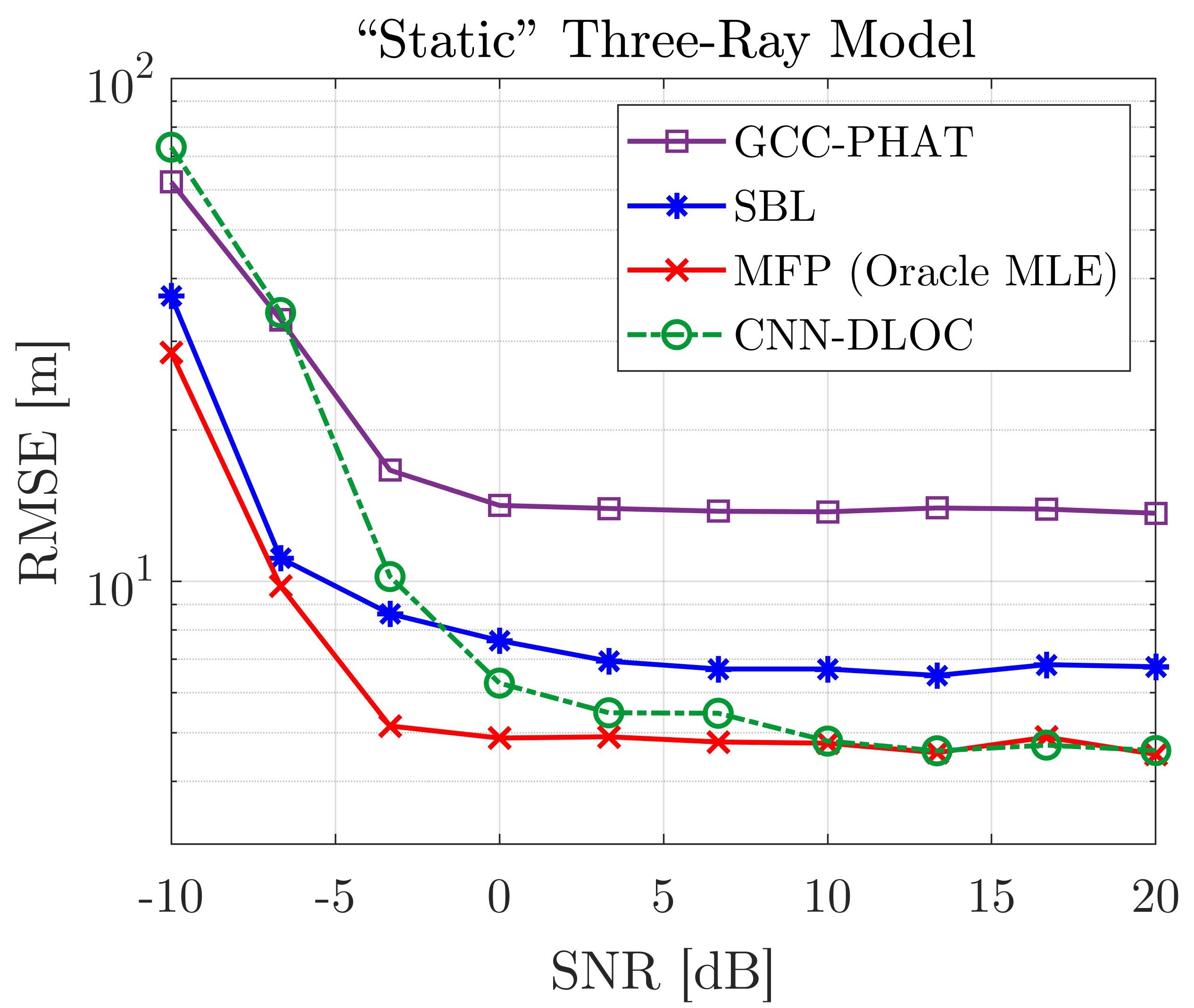}
	\caption{RMSE vs.\ SNR for the ``static" model, with a flat and stationary surface. Already from $0$ dB, CNN-DLOC dominates the non-oracle methods.}
	\label{fig:RMSEstatic}
\end{figure}

We first test our model with additive white CN noise, and compare its performance with three methods in the literature: the ``generalized cross-correlation with phase transform" (GCC-PHAT) \cite{zhang2008does} TDOA method, which is recognized as being resilient to multipath; SBL; and an oracle MFP, which uses the \emph{exact} values of $\{\ub_{\ell}\}$ for each realization. Fig.~\ref{fig:RMSEstatic} shows the root mean squared error (RMSE), $\sqrt{\Eset[\|\hup-\up\|_2^2]}$, vs.\ the SNR for each method.\footnote{Results are based on $10^3$ and $10^4$ trials for the model-based methods and CNN-DLOC, respectively, due to their different computational complexity.} As expected, the oracle MFP, which is also the oracle MLE in this case, dominates all methods. However, from $0$ dB onward our CNN-based DLOC model is the best non-oracle method, and at high SNR approximately \emph{matches the oracle performance}. We conclude that the model has the capacity to learn the statistics of the channel and the optimal estimation rule at high SNR.

\begin{figure}[h]
	\includegraphics[width=0.95\linewidth]{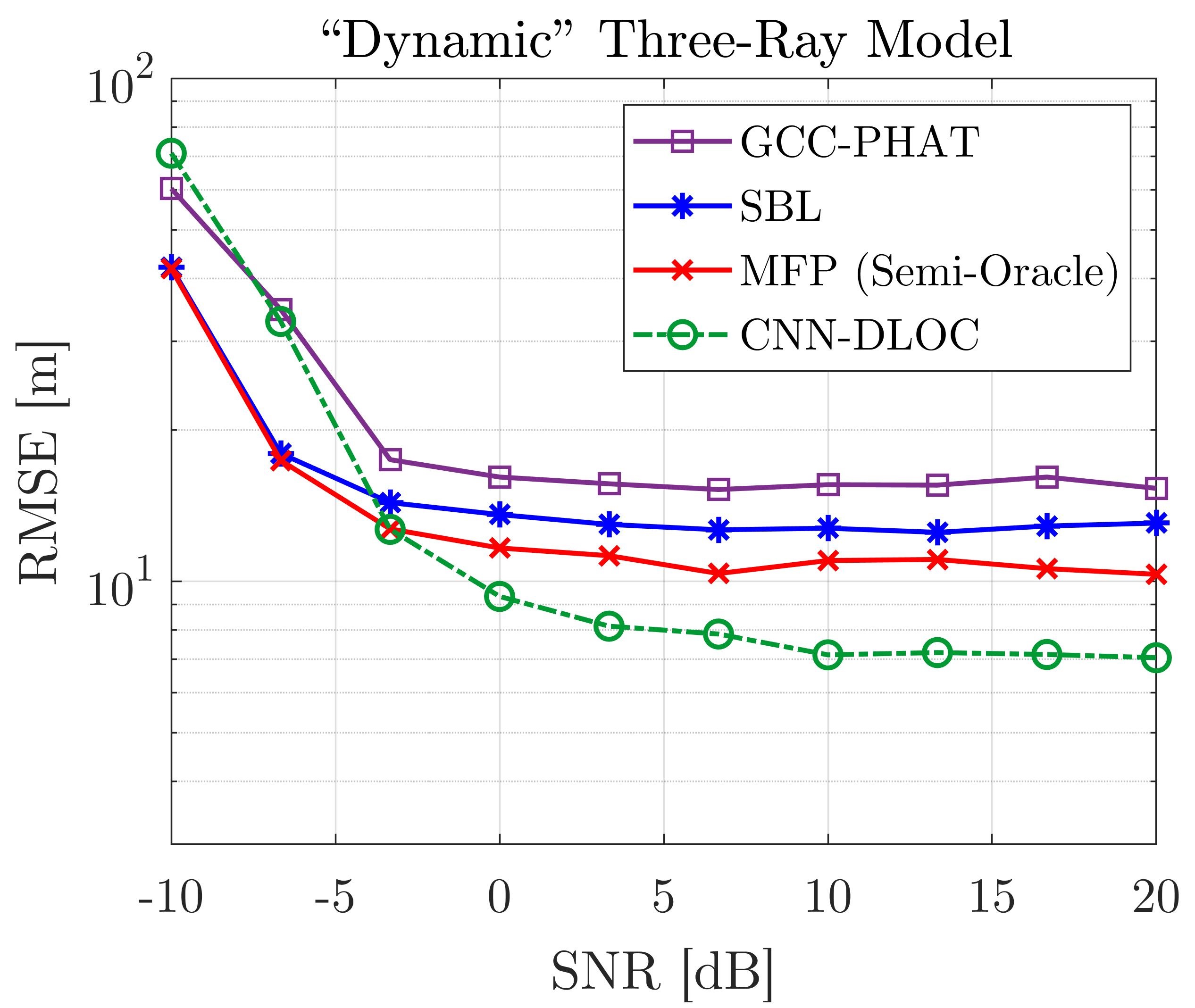}
	\caption{RMSE vs.\ SNR for the ``dynamic" model, with a flat, randomly shifted surface. Here, MFP is not the MLE, hence referred to as ``semi-oracle".}
	\label{fig:RMSEdynamic}
\end{figure}

\begin{figure}[h]
	\includegraphics[width=0.95\linewidth]{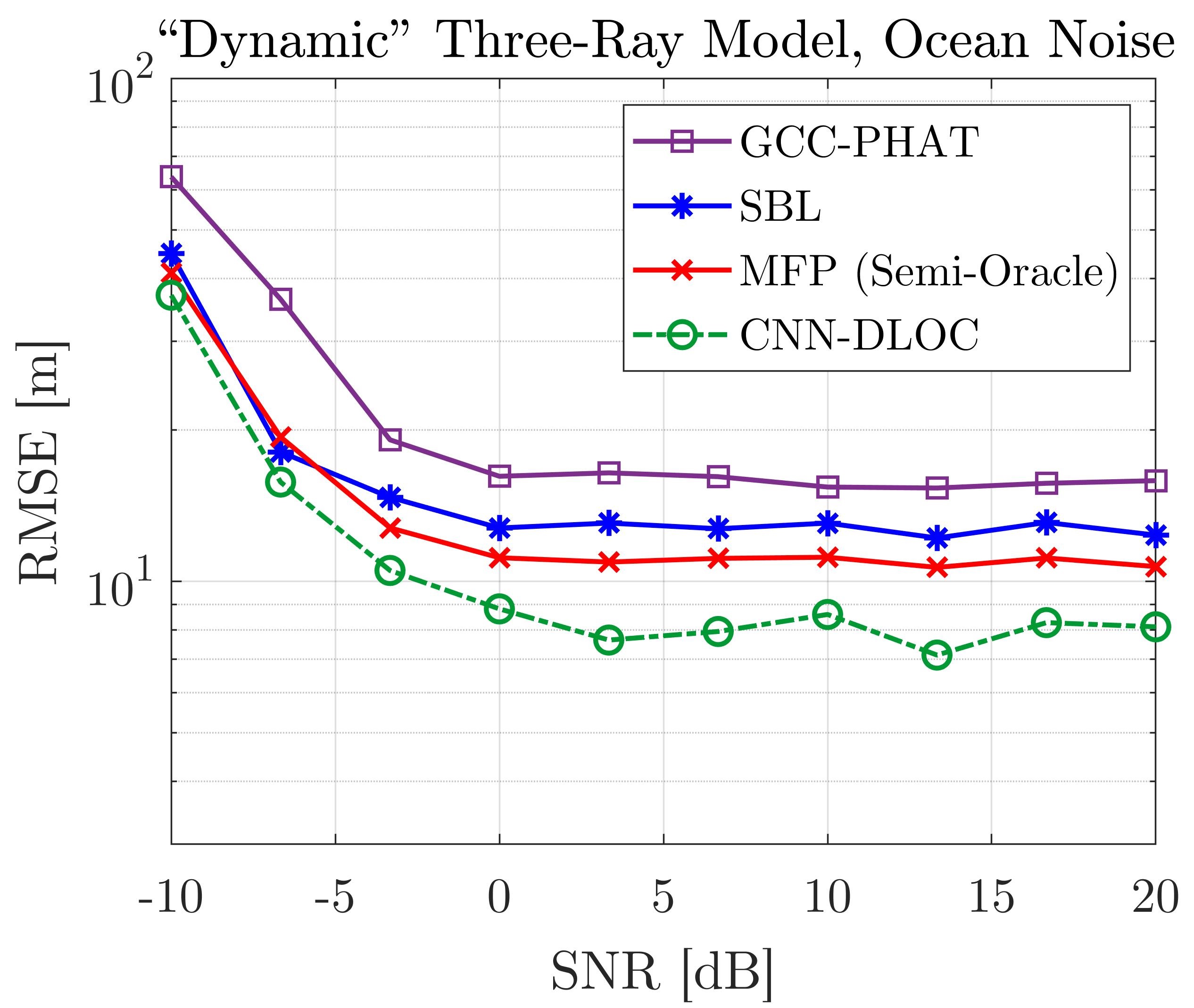}
	\caption{RMSE vs.\ SNR for the ``dynamic" model with ocean ambient noise. Here, CCN-DLOC dominates all methods, even at low SNRs.}
	\label{fig:RMSEdynamicKAM11}
\end{figure}

We repeat the experiment, for synthetic CN and real ocean ambient noise from the KAM11 experiment \cite{hodgkiss2012kauai}, but now for a ``dynamic" propagation model, in which the surface height is slightly perturbed, giving rise to perturbations in the surface-reflection associated time-delays, $\{\tau_{2\ell}(\up,\mathcal{E})\}$. To this end, in each realization we add zero-mean Gaussian perturbations to the time-delays $\{\tau_{2\ell}(\up,\mathcal{E})\}$, with a standard deviation of $1\%$ of the average distance from the source to the receivers, divided by $c$. Figs.~\ref{fig:RMSEdynamic} and \ref{fig:RMSEdynamicKAM11} reflect that due to these perturbations, the performances of the model-based methods already deteriorate considerably. In contrast, the proposed data-driven method, which was trained with data reflecting this phenomenon, adjusts and thereby attains the highest accuracy---from $0$ dB onwards for (synthetic) CN noise, and uniformly for real ocean ambient noise. These results demonstrate the potential to implement data-driven DLOC methods that go beyond what is offered by model-based methods derived for relatively simple propagation models. Further, such methods can learn the true underlying statistics of ocean ambient noise, and exploit them for higher localization accuracy.


\vspace{-0.25cm}
\section{Concluding Remarks}\label{sec:conclusion}
\vspace{-0.25cm}
We present a computationally efficient, data-driven direct localization approach, that can asymptotically match optimal model-based solutions, which are hard to implement in practice. The proposed holistic solution includes a specifically tailored architecture and a progressive training procedure. Still, many interesting aspects for potential research avenues remain to be explored. Examples include extensions to more complex environments, tracking, coarsely-quantized signals, scaling of the architecture (width and number of layers) with the volume of operation and more.

% \appendix
% \section{Alternate Form of the SBL Objective}\label{app:sblefficientcomp}
% Define the Cholesky decompositions \cite{golub2013matrix} 
% \begin{equation*}%\label{CholeskyoldAppendix}
% \D_{\ell}^{\tps}\D^*_{\ell}\triangleq\mGamma_{\ell}^{\her}\mGamma_{\ell}\in\Cset^{R\times R}, \; \forall \ell\in\{1,\ldots,L\},
% \end{equation*}
% where $\mGamma_{\ell}\in\Cset^{R\times R}$, and further define\begin{equation*}%\label{Umatrices}
% \U(\up,\mathcal{E})\triangleq\left[\bX_{1}\D^*_{1}\mGamma_{1}^{-1} \cdots \; \bX_{L}\D^*_{L}\mGamma_{L}^{-1}\right]\in\Cset^{N\times RL}.
% \end{equation*}
% Based on \cite[Prop.~3]{weiss2021semi}, $\lambda_{\max}\left(\Q(\up,\mathcal{E})\right)=\lambda_{\max}\left(\widetilde{\Q}(\up,\mathcal{E})\right)$, where $\widetilde{\Q}(\up,\mathcal{E})\hspace{-0.05cm}\triangleq\hspace{-0.05cm}\U(\up,\mathcal{E})^{\her}\U(\up,\mathcal{E})$, which, when written explicitly with $\G_{\ell}\hspace{-0.05cm}\triangleq\hspace{-0.05cm}\D^*_{\ell}\mGamma_{\ell}^{-1}$, gives the required alternate form.

% References should be produced using the bibtex program from suitable
% BiBTeX files (here: strings, refs, manuals). The IEEEbib.bst bibliography
% style file from IEEE produces unsorted bibliography list.
% -------------------------------------------------------------------------
% {\small \bibliographystyle{IEEEbib}
% \bibliography{strings,refs}}
\vspace{-0.2cm}
\bibliographystyle{IEEEtran}
\small{\bibliography{refs}}

% Generated by IEEEtran.bst, version: 1.14 (2015/08/26)
\begin{thebibliography}{10}
\providecommand{\url}[1]{#1}
\csname url@samestyle\endcsname
\providecommand{\newblock}{\relax}
\providecommand{\bibinfo}[2]{#2}
\providecommand{\BIBentrySTDinterwordspacing}{\spaceskip=0pt\relax}
\providecommand{\BIBentryALTinterwordstretchfactor}{4}
\providecommand{\BIBentryALTinterwordspacing}{\spaceskip=\fontdimen2\font plus
\BIBentryALTinterwordstretchfactor\fontdimen3\font minus
  \fontdimen4\font\relax}
\providecommand{\BIBforeignlanguage}[2]{{%
\expandafter\ifx\csname l@#1\endcsname\relax
\typeout{** WARNING: IEEEtran.bst: No hyphenation pattern has been}%
\typeout{** loaded for the language `#1'. Using the pattern for}%
\typeout{** the default language instead.}%
\else
\language=\csname l@#1\endcsname
\fi
#2}}
\providecommand{\BIBdecl}{\relax}
\BIBdecl

\bibitem{waterston2019ocean}
J.~Waterston, J.~Rhea, S.~Peterson, L.~Bolick, J.~Ayers, and J.~Ellen, ``Ocean
  of things: Affordable maritime sensors with scalable analysis,'' in
  \emph{Proc.\ OCEANS Conf.}, 2019, pp. 1--6.

\bibitem{tan2011survey}
H.-P. Tan, R.~Diamant, W.~K.~G. Seah, and M.~Waldmeyer, ``A survey of
  techniques and challenges in underwater localization,'' \emph{Ocean Eng.},
  vol.~38, no. 14-15, pp. 1663--1676, 2011.

\bibitem{bianco2019machine}
M.~J. Bianco, P.~Gerstoft, J.~Traer, E.~Ozanich, M.~A. Roch, S.~Gannot, and
  C.-A. Deledalle, ``Machine learning in acoustics: Theory and applications,''
  \emph{J.\ Acoust.\ Soc.\ Am.}, vol. 146, no.~5, pp. 3590--3628, 2019.

\bibitem{niu2019deep}
H.~Niu, Z.~Gong, E.~Ozanich, P.~Gerstoft, H.~Wang, and Z.~Li, ``Deep-learning
  source localization using multi-frequency magnitude-only data,'' \emph{J.\
  Acoust.\ Soc.\ Am.}, vol. 146, no.~1, pp. 211--222, 2019.

\bibitem{testolin2019underwater}
A.~Testolin and R.~Diamant, ``Underwater acoustic detection and localization
  with a convolutional denoising autoencoder,'' in \emph{Proc. of CAMSAP},
  2019, pp. 281--285.

\bibitem{gong2020machine}
Z.~Gong, C.~Li, and F.~Jiang, ``A machine learning-based approach for
  auto-detection and localization of targets in underwater acoustic array
  networks,'' \emph{{IEEE} Trans. Veh. Technol.}, vol.~69, no.~12, pp.
  15\,857--15\,866, 2020.

\bibitem{chen2021model}
R.~Chen and H.~Schmidt, ``Model-based convolutional neural network approach to
  underwater source-range estimation,'' \emph{J.\ Acoust.\ Soc.\ Am.}, vol.
  149, no.~1, pp. 405--420, 2021.

\bibitem{lefort2017direct}
R.~Lefort, G.~Real, and A.~Dr{\'e}meau, ``Direct regressions for underwater
  acoustic source localization in fluctuating oceans,'' \emph{Applied
  Acoustics}, vol. 116, pp. 303--310, 2017.

\bibitem{wang2018underwater}
Y.~Wang and H.~Peng, ``Underwater acoustic source localization using
  generalized regression neural network,'' \emph{J.\ Acoust.\ Soc.\ Am.}, vol.
  143, no.~4, pp. 2321--2331, 2018.

\bibitem{qin2020underwater}
D.~Qin, J.~Tang, and Z.~Yan, ``Underwater acoustic source localization using
  {LSTM} neural network,'' in \emph{2020 39th Chinese Control Conference
  (CCC)}, 2020, pp. 7452--7457.

\bibitem{weiss2004direct}
A.~J. Weiss, ``Direct position determination of narrowband radio frequency
  transmitters,'' \emph{{IEEE} Signal Process. Lett.}, vol.~11, no.~5, pp.
  513--516, 2004.

\bibitem{wang2020direct}
L.~Wang, Y.~Yang, and X.~Liu, ``A direct position determination approach for
  underwater acoustic sensor networks,'' \emph{{IEEE} Trans. Veh. Technol.},
  vol.~69, no.~11, pp. 13\,033--13\,044, 2020.

\bibitem{weiss2021semi}
A.~Weiss, T.~Arikan, H.~Vishnu, G.~B. Deane, A.~C. Singer, and G.~W. Wornell,
  ``A semi-blind method for localization of underwater acoustic sources,''
  \emph{IEEE Trans. Signal Process.}, vol.~70, pp. 3090--3106, 2022.

\bibitem{etter2018underwater}
P.~C. Etter, \emph{Underwater acoustic modeling and simulation}.\hskip 1em plus
  0.5em minus 0.4em\relax CRC Press, 2018.

\bibitem{porter2011bellhop}
M.~B. Porter, ``The {BELLHOP} manual and user’s guide: Preliminary draft,''
  \emph{Heat, Light, and Sound Research, Inc., La Jolla, CA, USA, Tech. Rep},
  vol. 260, 2011.

\bibitem{hodgkiss2012kauai}
W.~Hodgkiss and J.~Preisig, ``Kauai {ACOMMS} {MURI} 2011 ({KAM}11)
  experiment,'' in \emph{Proc. of Eur. Conf. Underwater Acoust.}, 2012, pp.
  993--1000.

\bibitem{menze2017influence}
S.~Menze, D.~P. Zitterbart, I.~van Opzeeland, and O.~Boebel, ``The influence of
  sea ice, wind speed and marine mammals on southern ocean ambient sound,''
  \emph{Royal Society Open Science}, vol.~4, no.~1, p. 160370, 2017.

\bibitem{baggeroer1993overview}
A.~B. Baggeroer, W.~A. Kuperman, and P.~N. Mikhalevsky, ``An overview of
  matched field methods in ocean acoustics,'' \emph{IEEE J. Ocean. Eng.},
  vol.~18, no.~4, pp. 401--424, 1993.

\bibitem{weiss2022mlspsuppmat}
A.~Weiss, T.~Arikan, and G.~W. Wornell, ``Blind direct localization via
  convolutional neural networks: A data-driven approach, supplementary
  materials,'' \url{https://www.weissamir.com/project/DLOC}, 2022.

\bibitem{stojanovic2009underwater}
M.~Stojanovic and J.~Preisig, ``Underwater acoustic communication channels:
  Propagation models and statistical characterization,'' \emph{IEEE Commun.
  Mag.}, vol.~47, no.~1, pp. 84--89, 2009.

\bibitem{zhang2008does}
C.~Zhang, D.~Flor{\^e}ncio, and Z.~Zhang, ``Why does {PHAT} work well in low
  noise, reverberative environments?'' in \emph{Proc. of ICASSP}, 2008, pp.
  2565--2568.

\end{thebibliography}

\end{document}